\documentclass{article}
\usepackage{ijcai19}

\usepackage{times}
\usepackage{soul}
\usepackage{url}
\usepackage[hidelinks]{hyperref}
\usepackage[utf8]{inputenc}
\usepackage[small]{caption}
\usepackage{graphicx}
\usepackage{amsmath}
\usepackage{booktabs}
\usepackage{algorithm}
\usepackage{algorithmic}
\title{A Survey on Neural Machine Reading Comprehension}

\author{
Boyu Qiu $^\dag$ \footnote{Co-first authours. B.Qiu. and X.Chen. contributed equally.} \and 
Xu Chen $^\dag$ $^*$  \and
Jungang Xu \footnote{Corresponding authours.}  \and 
Yingfei Sun
\affiliations
University of Chinese Academy of Sciences
\emails
\{qiuboyu18, chenxu18\}@mails.ucas.ac.cn \\
\{xujg, yfsun\}@ucas.ac.cn
}

\begin{document}

\maketitle

\begin{abstract}
  Enabling a machine to read and comprehend the natural language documents so that it can answer some question remains an elusive challenge. In recent years, the popularity of deep learning and the establishment of large-scale datasets have both promoted the prosperity of Machine Reading Comprehension. This paper aims to present how to utilize the Neural Network to build a Reader and introduce some classic models, analyze what improvements they make. Further, we also point out the defects of existing models and future research directions.
\end{abstract}

\section{Introduction}

As an important mean of measuring the machine's ability to understand documents, Machine Reading Comprehension (MRC) has been the central goal of Natural Language Processing (NLP) research . MRC tasks usually require the machine to answer questions after reading one or more relevant passages. Traditional methods tend to use pattern matching techniques with additional automated linguistic processing like stemming, name identification, or analyzing the relationship between questions and answers, then use statistical or mathematical methods to calculate the similarity between the question and sentences to find the most likely answer. Such methods are more like based on retrieval of the relational database rather than the real understanding of given passages, so they lack of reasoning ability.

With the popularity of deep learning and the emergence of some brilliant architecture in NLP such as the sequence-to-sequence model and word vectors, MRC ushered in the stage of prosperity in 2015. In this year, \cite{hermann_teaching_2015} proposed two basic neural MRC models: attentive reader and impatient reader, which provide a lot of heuristic ideas, such as using bidirectional RNN/LSTM to capture the contextual semantics of the passage, using the attention mechanism to make the machine imitate human to read with questions. These ideas have become the basis of many later neural MRC models. In 2016, Stanford University released the Stanford Question Answering Dataset (SQuAD) \cite{rajpurkar_squad:_2016}, which contains more than 100,000 (questions, original, answer) triples. After that, a large number of representative models are proposed on this dataset. Large-scale datasets and end-to-end neural network models have jointly promoted the prosperity of MRC. Recent work such as GPT \cite{radford_improving_nodate}, BERT \cite{devlin_bert:_2018}, etc. have surpassed human performance in several datasets.

The purpose of this paper is to analyze and summarize the classic models of neural MRC in recent years. In Section 2, we divide the existing dataset into four categories based on the question style and introduce them separately. In Section 3, we first extract the common architecture of existing models. Then for some efficient models, we present improvements what they make briefly. Besides, we also introduce how to utilize transfer learning to solve NLP tasks. In Section 4, we provide the application of machine reading comprehension, including conversational question answering and open-domain question answering, etc. The Sction 5 is discussion, presenting some experiments from the recent achievements and some inadequacies of the current MRC tasks.

\section{Datasets}

In the exploration of MRC, constructing a high-quality, large-scale dataset is as important as optimizing the model structure. According to the form of question and answer, the MRC datasets can be roughly divided into four categories. 

\begin{table*}
\centering
\begin{tabular}{lllrrl}
\toprule
Dataset &Answer Type &Data Source &Question &Document  &Metric\\
\midrule
CNN/DM &Cloze-style &News report &1.4M &300k &Accuracy\\
CBT &Cloze-style &Children's book &688k &108 &Accuracy\\
MCTest &Multi-choice &Fictional stories &2k &500 &Accuracy\\
RACE &Multi-choice &English exams &97k &28k &Accuracy\\
SQuAD &Extractive &Wikipedia &100k &536 &EM, F1\\
SQuAD2.0 &Extractive, Unanswerable &Wikipedia &150k &505 &EM, F1\\
TriviaQA &Extractive &Wikipedia,Web &40k &660k &EM, F1\\
NarrativeQA &Abstractive &Wikipedia &46k &1.5K &ROUGE-L\\
DuReader &Abstractive, Unanswerable &User logs &200k &1M &BLEU, ROUGE-L\\
MC MARCO &Abstractive, Unanswerable &User logs &1M &3.2M &Accuracy, ROUGE-L, BLEU\\
\bottomrule
\end{tabular}

\caption{Datasets}
\label{1}
\end{table*}

\subsection{Cloze-Style}

In this style MRC tasks, the question contains a placeholder and the machine must decide which word or entity is the most suitable option.During evaluation, accuracy is the only metric. Main datasets in this style are listed below.

\paragraph{CNN/Daily Mail:}\cite{hermann_teaching_2015} collected 93k articles from the CNN and 220k articles form the Daily Mail (DM). The dataset extracts the entity of the critical sentence in the news story and replaces it with a placeholder. To remove the influence of external word knowledge, the named entities in the dataset are replaced by anonymous IDs, which are then further shuffled for each example. This processing makes the models to rely on text to answer questions.

\paragraph{Children's Book Test:}\cite{hill_goldilocks_2015} proposed Children's Book Test dataset (CBT) using a different style. They take a sequence of 21 consecutive sentences from a children's book. Then, the first 20 sentences are viewed as context and the query is to infer the missing word in the 21st sentence.

\subsection{Multi-Choice}

This style MRC tasks require the machine to find the only correct option in the k (usually 4) hypothesis based on the given passage. Similarly, the evaluation metric is accuracy. Main datasets in this style are listed below.

\paragraph{MCTest:}\cite{richardson_mctest:_nodate} constructed MCTest, the first comprehensive reading comprehension dataset since the wave of neural networks, which has 660 fictional stories, each with 4 questions and 4 candidate answers.

\paragraph{RACE:}\cite{lai_race:_2017} collected more than 20,000 articles and 100,000 questions from Chinese middle and high school students’ English exams, covering a wide range of fields. These questions were proposed by experts, which was orig-inally to examine the level of human reading comprehension. Therefore, answering the question requires the machine to have some reasoning ability.

\subsection{Span-Prediction}

The machine needs to find the correct answer from one of the single spans in the passage. This category is also referred to as {\it extractive question answering}. There are two metrics to evaluate the task. {\bf EM} (Exact Match) assigns credit 1.0 if the predicted answer is equal to the standard answer and 0.0 otherwise. {\bf F1} computes the average word overlap between the predicted and the standard answer. Main datasets in this style are listed below.

\paragraph{SQuAD:}\cite{rajpurkar_squad:_2016} proposed the reading comprehension dataset SQuAD, which contains more than 100,000 questions identified in 536 Wikipedia by manual workers. Each question corresponds to a specific passage, and the answer to the question is located at a span of the passage. The challenge based on the SQuAD greatly promoted the prosperity of MRC. 

\paragraph{SQuAD2.0:}In 2018, the SQuAD 2.0 version was released by \cite{rajpurkar_know_2018}. Unlike the previous version, more than 50,000 new, unanswerable questions have been added. These questions have plausible answers in the corresponding article, that is, the questions are of the same type, but not correct. The model needs to identify these questions to avoid answering them. 

\paragraph{TriviaQA:}\cite{joshi_triviaqa:_2017} TriviaQA, which contains over 650K question-answer-evidence triples.In comparison to the other datasets, TriviaQA has relatively considerable syntactic and lexical variability between questions and corresponding answer-evidence sentences and requires more cross sentence reasoning to find answers.

\subsection{Free-Form}

The last category allows the answer to be any free-text form. Because the standard answer is human-generated, there isn’t a consensus yet on the evaluation metric. A common way is to use standard evaluation metrics in natural language generation tasks such as BLEU (the metric of {\it machine translate}), GOUGE (the metric of {\it automatic abstract}).Main datasets in this style are listed below.

\paragraph{NarrativeQA:}\cite{kocisky_narrativeqa_2017} proposed NarrativeQA, a more difficult data set, aimed to increase the difficulty of the question, making it not easy to locate the answer. The dataset contains 1567 complete stories from books and screenplays. Questions and answers are written by humans and are mostly more complex styles such as "when/where/who/why."

\paragraph{Dureader:}\cite{he_dureader:_2017} proposed DuReader, a Chinese reading comprehension dataset, which collected more than one mil-lion documents and more than 200,000 different types of questions in Baidu Zhidao and Baidu Search, including Yes/No questions, descriptive questions, etc(up to 2018.6). 

\paragraph{MS MACRO:}\cite{bajaj_ms_2016} introduced a large- scale machine reading comprehension dataset, which comprises of 8,841,823 pas-sages and 1,010,916 anonymized questions sampled from Bing’s search query logs(up to 2018.10).A question in the MS MARCO dataset may have multiple answers or no answers at all.

\paragraph{}In addition, there have been some new forms of datasets in recent years, such as {\bf CoQA}: a conversational QA dataset proposed by \cite{reddy_coqa:_2018}. Each conversation has an average of 15 rounds. The answer to each round of dialogue is based on the questions in the article and the previous rounds of dialogue, and the answer is open-form.

\begin{figure}
	\centering
	\includegraphics[width=8cm]{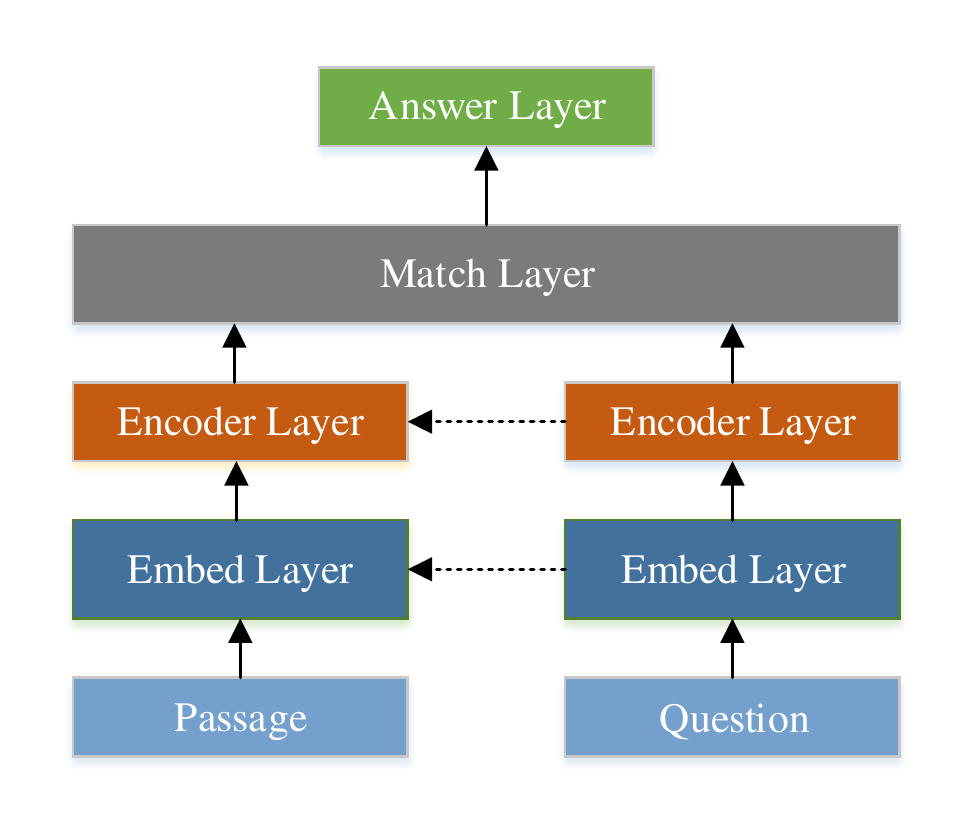}
	\caption{A generic architecture of MRC model}
	\label{architrcture}
\end{figure}

\section{Models}

\subsection{Generic Architecture}

We have extracted a generic architecture of existing MRC models. This architecture, which we can see as an extension of the Encoder-Decoder model proposed by \cite{sutskever_sequence_2014}, is shown in Figure \ref{architrcture}. The original passage and the question are represented by the word vector through the Embed layer and the word vector is encoded by Encoder Layer to aggregate the context semantic information. The match layer is used to capture the interaction between passage and question, and a query-aware representation of the original passage is then obtained. The common idea here is attention mechanism, that is, the machine is better to pay more attention to the part of the passage that is most related to the question, which is the critical improvement part of most models. The Answer Layer is based on the passage representation from the match layer and predicts the answer, while the way of implementation varies according to the type of question.

\subsubsection{Embed Layer}

Word embedding, a pre-training process for NLP tasks, can project one-hot word representation into low dimensional space. In one-hot feature space, the dimension is as large as the size of vocabulary size ($V$), which always incurs the {\it dimension curse}. While in the embedded space, the dimension ($N$) can be arbitrary (usually 256, 512). In addition, word embedding can capture some semantic features between words, such as similarity, analogy.

What the word embedding actually does is to find an embedded matrix, whose dimension is $V\times N$.There are three classic methods to do that. {\bf CBOW} predicts a word using the context word of the passage, while {\bf Skim-Gram} predicts context words of the passage using a given word. Both of them ({\it word2vec}) were proposed by \cite{mikolov_efficient_2013}.In addition, \cite{pennington_glove:_2014} proposed {\bf Glove}, using global word frequency statistics to construct the embedded matrix. Some models not only use the word embedding, but also combine the {\it character embedding}, which is acquired by the character-level Convolution Neural Network (CNN) or Recurrent Neural Network (RNN), to process the corpus. 

However, in a particular context, a word has only one semantic. The existence of polysemous words makes the word embedding a mixture of multiple semantics, which limits the representation performance of word embedding. To tackle the puzzle, \cite{peters_deep_2018} proposed {\bf ELMo} (Embedding from Language Models), a dynamic word representation model. In the training process, ELMO not only learns a word embedding but also gains a two-layer, bidirectional LSTM structure. We can regard them as three semantically different word representations. When a word appears in the context, the weights of its three word representations are dynamically adjusted, and the word embedding generates by weighted summation. 

Using pre-trained word embedding can improve the efficiency of training, some models use Glove (\cite{seo_bidirectional_2016}) or word2vec (\cite{MEMEN}) in embed layer. ELMo is also heavily used in the MRC task (\cite{hu_reinforced_2017}).

\subsubsection{Encoder Layer}

After the Embed layer, the embedded token in the passage and the question is fed to encoder Layer for modeling{} the sentence and passage by RNN, which is widely used in NLP because of its excellent performance in time and sequence feature tasks. Uni-directional RNN can capture the context information on the left side of a sequence, while bidirectional RNN can aggregate the context information on both the left and right sides of a sequence. However, RNN cannot capture long-distance dependencies, which is  essential to comprehend the passage. Hence, two variants of RNN, Long Short Memory Network (LSTM) and Gated Recurrent Unit (GRU), with their bidirectional versions were proposed. The early models use RNN, while the recent models tend to use LSTM (\cite{hermann_teaching_2015}) or GRU (\cite{R-net}).

\subsubsection{Match Layer}

At present, most models choose to use the attention mechanism (or its variants) in the match layer, which is very suitable to mine the association between passage and question. The essence of the attention mechanism can be described as a mapping process of the {\it query} to a series of {\it key-value} pairs (in NLP tasks, {\it key = value}). This process first use a function $f$ to measure the relevance between {\it query} and {\i{}t key} to gain the weight of {\it value} (Equation \ref{attention1} gives four common calculation methods), then normalize these weights using a softmax function(Equation \ref{attention2}). The values and the corresponding weights are summed to get the final attention (Equation \ref{attention3}).

\begin{equation}
	f(Q,K_{i})=\left\{
		\begin{aligned}
		&Q^TK_i &&dot\\  
		&Q^TW_aK_i &&general\\
		&W_{\alpha}[Q,K_{i}] &&concat\\
		&v_{\alpha}tanh(W_{\alpha}Q+U_{\alpha}K_{i}) &&perpceptron
		\end{aligned}
	\right.
	\label{attention1}
\end{equation}

\begin{equation}
	a_{i}=softmax(f(Q,K_{i}))=\frac{exp(f(Q,K_{i}))}{\Sigma_{j}exp(f(Q,K_{j}))}
	\label{attention2}
\end{equation}

\begin{equation}
	Attention(Q,K,V)=\Sigma_{i}a_{i}V_{i}
	\label{attention3}
\end{equation}

\subsubsection{Answer Layer}

Before entering the answer layer, the model has obtained the query-aware representation of passage. On this basis, the answer layer has different implementation methods to solve various tasks.

For the {\it cloze-style} question, the Reader calculate the similarity between every candidate answer with the passage, and choose the largest one. The conventional way of solving {\it multi-choice} question is almost the same as what cloze-style do: each option is encoded with LSTM or GRU, then compared with the passage, and choose the most likely answer. 

For the {\it span-prediction} question, {\bf Pointer Networks} \cite{vinyals_pointer_2015} are quite suitable to determine the answer. \cite{wang_machine_2016} proposed two ways of using Pointer Net. One is the sequence model, which predicts all the words in the sentence, and the other is the boundary model, which only predicts the start and end points of the answer. To rank the candidate spans, feed-forward neural networks are always used to score the similarity. 

{\it Free-form} question is more difficult to handle than the previous ones because it does not limit the form of the answer. The usual processing method is the same as the sequence-to-sequence model, and LSTM is used to decode the passage of query-aware to get the final answer.

\subsection{Classic Model}

Many classic models are based on the previous architecture. Specially, \cite{hermann_teaching_2015} proposed two neural MRC models: AR (Attentive Reader) and IR (Impatient Reader), which are instructive for later researches. For example, using BiLSTM to encode the corpus and using attention mechanism in match layer, have been  choices of most models. AR and IR also represent the two basic structures of the existing model and we will introduce them below.

\subsubsection{Two Basic Structures}

AR encodes the query using bidirectional LSTM and concatenates the results to get the representation of the query (Formula \ref{AR1}), which contains both forward semantics and reverse semantics. For the document, the composite output for each token at position t is Formula \ref{AR2}. Further, the representation of the document is formed by a weighted sum of these output vectors. The model is completed with the definition of the joint document and query embedding via a nonlinear combination (Formula \ref{AR3}):

\begin{equation}
	u = \overrightarrow{y_{q}}(|q|)||\overleftarrow{y_{q}}(1)
	\label{AR1}
\end{equation}

\begin{equation}
	y_{d}(t)=\overrightarrow{y_{d}(t)}||\overleftarrow{y_d(t)}
	\label{AR2}
\end{equation}

\begin{equation}
	g^{AR}(d,q)=tanh(W_{rg}r+W_{ug}u)
	\label{AR3}
\end{equation}

The improvement of IR lies in the encoding of the query. Similar to document, the model computes a query representation (Formula \ref{IR1}) at each token i of the query. The result is an attention mechanism that allows the model to recurrently accumulate information from the document as it sees each query token, ultimately outputting a final joint document query representation for the answer prediction(Formula \ref{IR2}).

\begin{equation}
	u = \overrightarrow{y_{q}}(|i|)||\overleftarrow{y_{q}}(i)
	\label{IR1}
\end{equation}

\begin{equation}
	g^{IR}(d,q)=tanh(W_{rg}r(|q|)+W_{qg}u)
	\label{IR2}
\end{equation}

\begin{figure}
	\centering
	\includegraphics[width=8cm]{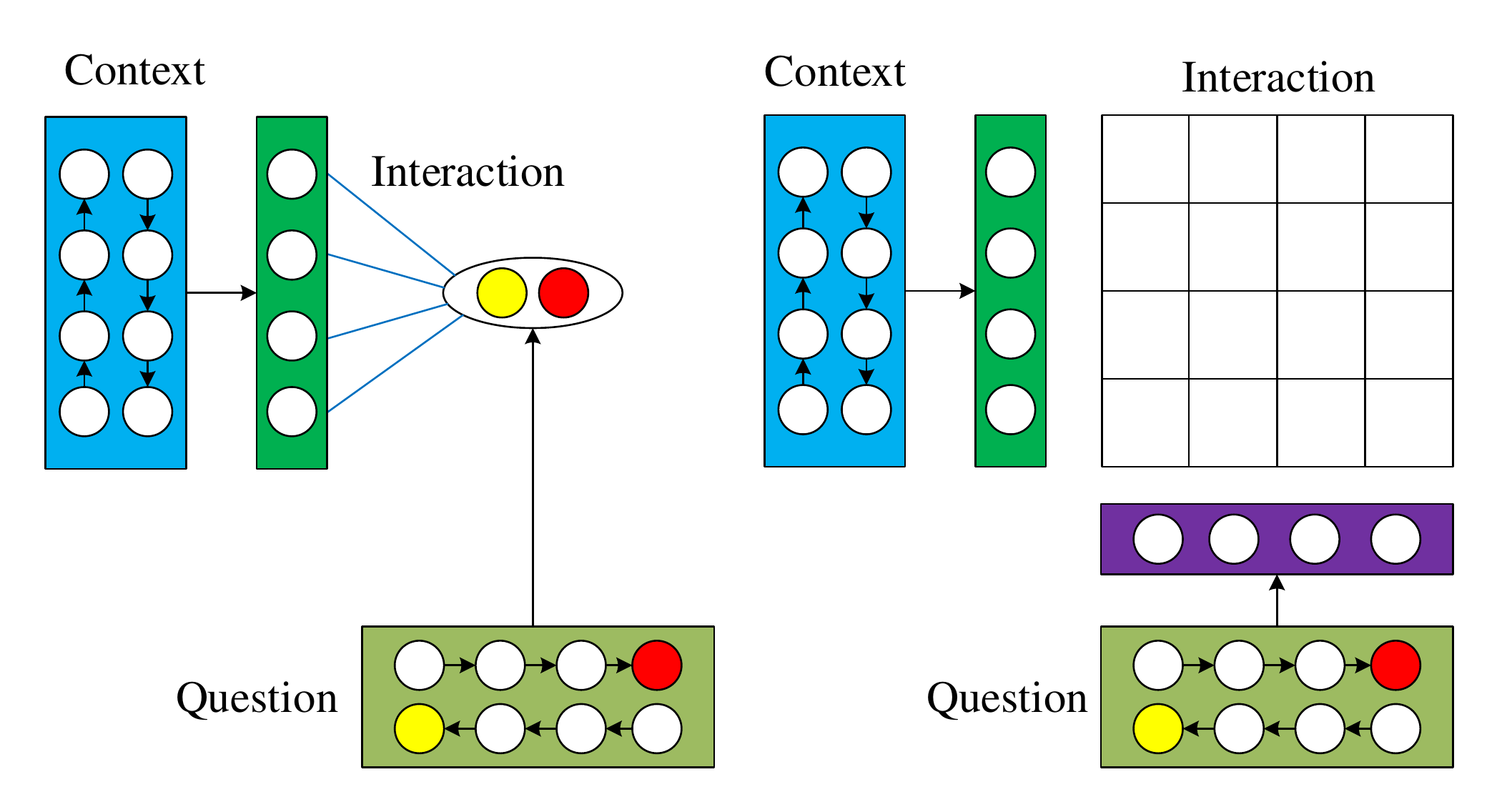}
	\caption{The left is the process of AR and the right is IR. In AR, the match layer, or the interaction(fusion) layer, is a vector. While in IR, a matrix W is obtained by the standard word-by-word attention mechanism, where $W_{ij}$ indicates the degree of matching between the i-th token in the passage and the j-th token in the question.}
	\label{IR and AR}
\end{figure}

According to Figure \ref{IR and AR}, we can see that the fundamental difference between AR and IR is that the representation of the query is based on the single token or the whole sentence.

\subsubsection{Heuristic Improvements}

Afterward, many excellent models appeared to improve or adjust the match layer of AR and IR. For example, the {\bf Attention Sum Reader} (Kadlec et al. 2016), uses dot produce between the question embedding and the contextual embedding to select the answer. The {\bf Stanford Attentive Reader} \cite{chen_thorough_2016} uses a bilinear term, instead of a {\it tanh} layer to compute the relevance between question and contextual embeddings. Both models are the variants of AR.

Intuitively, IR can introduce more detailed information of question, which is critical in MRC tasks. Indeed, the performance of the relevant models is better, such as the {\bf Match-LSTM Reader} \cite{wang_machine_2016}, the {\bf BiDAF}(Bi-Directional Attention flow) \cite{seo_bidirectional_2016}.

The Match-LSTM Reader also uses LSTM to preprocess the embedded passage and question. Then, the i-th row of the weight matrix is multiplied by the question representation, which means that the i-th word in the passage matches the question. Through the bidirectional Match-LSTM, the reader gets the overall semantics of the whole question. Further, the pointer net is use to look for the answer.

BiDAF is the first model to combine character embedding  with word embedding in the embed layer. In addition, the other improvement of BiDAF is reflected in the introduction of a bidirectional attention mechanism, that is, first calculating the alignment matrix of the original passage and the question, and then calculating the Query2Context attention and Context2Query attention based on the matrix to get the query-aware passage representation. Finally, the reader uses BiLSTM to aggregate context information.

\subsubsection{Innovation on Attention Mechanism}

The excellent performance of BiDAF prompted more exploration of attention mechanism. Several models for improving attention gradually appears and achieves better performance. For example, the new feature extractor {\bf Transformer}(see 3.3) \cite{vaswani_attention_2017}, {\bf DCN}(Dynamic Coattention Networks) \cite{xiong_dynamic_2016}, the {\bf FastQA} \cite{weissenborn_making_2017}, the {\bf R-net} and so on.

DCN uses a coattention mechanism in the Encoder part, which generates weight attention for both context and question using a bidirectional attention mechanism. Besides, to recover from the local optimum (incorrect solution), DCN proposed the Dynamic Pointing Decoder, which dynamic iteratively predicts the start and the end point. Specifically, the Highway Maxout Network scores each word in every document as the start point (or end point) based on the historical prediction information and the previous prediction situation. It stores the historical prediction information with LSTM and takes the position with the highest score in the last iteration as the start position (or end position).

FastQA uses a lightweight architecture with no interaction layer. In the Embed layer: in addition to word and char embedding as input, for each word in the context, two word-in-question features are used to replace the attention mechanism: one is the binary feature, indicating whether the word appears in the question; the other is the weighted feature, which combines the token's term frequency in the question with the inverse term frequency in the context. These two features are also used in the question words. In the answer Layer, FastQA uses {\it Beam-search} to determine the answer range. FastQAExt (the extended version of FastQA) has an interaction layer: two lightweight information fusion strategies are used: Intra-Fusion and Inter-Fusion, the former calculates the similarity between each word in the original text, and the latter calculates the similarity between each original word and the question word. Then send it to Highway Networks to get the query-aware passage representation.

R-NET introduces a self-attention mechanism into the model. The Reader chooses Glove and char-embedding to pre-train the question and passage and uses BiRNN to get representation answer. Then, a Gated attention-based recurrent network incorporates the question into the passage to reach the question-aware passage representation. Further, self-matching attention layer is used to capture the knowledge of the context. Ultimately, R-net uses pointer networks to predict the start and end position of the answer. In addition, an attention-pooling over the question representation is used to generate the initial hidden vector for the pointer network.

There are also many other brilliant innovationas, such as the forthputting of {\it reinforcement learning}. {\bf ReasoNet} does not have a fiexed approach of turns during inference, but utilizes {\it reinforcement learning} to dynamically determine whether to continue the comprehension results \cite{shen_reasonet:_2017}. The {\bf Mnemonic Reader} encourages to predict a more acceptable answer to address the {\it convergence suppresion} by dynamic-critical {\it reinforcement learning}. Besides, its reattention mechanism is effecive to avoid attention redundancy and attention deficency \cite{hu_reinforced_2017}.

\subsection{Transfer Learning Model}

All models have been improving performance by attempting to mine the information contained in the context. But there is a fact that when people do reading-comprehension tasks, they often use a lot of knowledge that does not appear in the context. For example, we always default everyone knows that Roosevelt is from the United States while Churchill is from the United Kingdom. For the phenomenon, transfer learning is a heuristic thought.

Recently, the pre-training model has achieved excellent results in NLP tasks. The basic structure of such a model is to conduct semi-supervised pre-training on the large corpus and supervised fine-tuning on specific tasks (the process is called transfer learning). These models achieved the best performance in several datasets. Up to now, successful pre-training models mainly include: {\bf GPT}, {\bf BERT}, and {\bf ELMO} (details in 3.1). Before introducing them, it is essential to know what the Transformer is. 

\subsubsection{Transformer}

Previously, RNN, LSTM or GRU is be seen as the best choice in sequence modeling or language models, but all these recurrent architectures have a fatal flaw, which is hard to parallel in the training process, limiting the computational efficiency. \cite{vaswani_attention_2017} proposed Transformer, based entirely on self-attention rather than RNN or Convolution.

The Encoder part consists of six identical network layers. Each layer has two sub-layers, the first is the multi-head self-attention mechanism, and the second is point-wise fully connected feed-forward network. Layer normalization and residual connection are applied to every sublayer. The decoder part has the almost same architecture with the encoder, except that it adds encoder-decoder attention layer between the self-attention layer and feed-forward neural network.

Transformer's most notable improvement is the application of self-attention and multi-headed attention: Self-attention, also known as Intra attention, captures attention between words in a sequence. The attention calculation method here uses Scaled Dot-Product Attention in formula \ref{self-attention}, where $d_{k}$ indicates the dimension of keys.

\begin{equation}
	Attention(Q,K,V)=softmax(\frac{QK^{T}}{\sqrt{d_{k}}})V
	\label{self-attention}
\end{equation}

Multi-headed attention means that the original sequence is mapped by several mapping functions to obtain several sets of Q, K, V. Each set of Q, K, V values can be used to compute attention in parallel and the resulting values are then concatenated. The advantage of this approach is that it extends the ability of the model to focus on different locations, giving multiple representation subspaces of the attention layer.

\subsubsection{GPT and BERT}
The transformer can not only achieve parallel calculations but also capture the semantic correlation of any span. Therefore, more and more language models tend to choose it to be the feature extractor.

{\it OpenAI} proposed a generative pre-training model based on the transformer, using a multi-layer transformer decoder for language modeling. Unlike the {\it vanilla} transformer, it does not contain the encoder-decoder attention layer. This model achieved an absolute improvement of 8.9\% on commonsense reasoning and optimal results on RACE dataset. However, {\it OpenAI} transformer has one glaring flaw: it only trains forward language models.

Unlike recent language representation models, BERT is designed to pre-train deep bidirectional representations by jointly conditioning on both left and right context in all layers. After training on the BooksCorpus (800M words) and Wik-ipedia (2,500M words) to learn some expressions of the language, BERT can achieve the best performance on SQuAD dataset with just one additional fine-tuned output layer. In addition to large amounts of data and calculations, BERT's superior performance is based on Masked Language Model (MLM) and Next Sentence Prediction (NSP) during the training process. The former is to capture semantics between words and words (similar to CBOW) and the latter is to discover the logic between sentences and sentences.

\section{Applications}

The application of MRC has two general directions: one is Conversational Question Answering, where a machine has to understand passages and answer questions that appear in a conversation, the other is Open-domain Question Answering, where a machine needs to answer questions without passage.

The conversational question answering is directly related to dialogue, which is the core of building dialogue systems to converse with humans in natural language. Usually, the dialogue systems are designed for a particular task and set up to have short conversations,(e.g.booking a flight). When dealing with such tasks, the Reader often has to consider the sequence of questions. For example, the first question is “How many flights to Paris today?”,  the second question is “Which is the earliest?”. \cite{guo_dialog--action:_nodate} gave a comprehensive survey on how to use deep neural networks to build dialogue systems.

Unlike the previous tasks, the open-domain question answering does not specify the passage of the given question. One solution is to build large-scale structured Knowledge Bases, such as DBpedia, and then retrieve the answers in the KB. This method is very intuitive, but because external knowledge is always growing, it's costly to construct and maintain KB. Another way is to combine Information Retrieval (IR) and MRC. This method is generally divided into two stages. First, the passages are sorted according to the degree of relevance to the question from the Knowledge Source (Wikipedia and so on), and the first {\it n} articles are taken as the base of the reading comprehension process. With these articles, the reading comprehension model mentioned earlier can be applied to get the answer. 

\section{Discussion}

Since 2015, MRC has made very significant achievements, which have attracted the attention of many researchers.

The construction of large-scale corpus is a necessary prerequisite for promoting the development of MRC. From the CNN/DM, to the recent SQuAD dataset, and then to the MS MARCO, DuReader, each new dataset has presented new problems which current methods cannot effectively solve . New issues have prompted researchers to continually explore new models and promote the development of the field.

The match layer built on attention mechanism is essential for the model to understand the original passage and the question. The more complex bidirectional interaction layer design, which enables more semantic information to flow in the original passage, thus partially solving the long-distance dependencies in the  passage, is undoubtedly better than the previous single interaction layer design. In particular, self-attention mechanism, which can capture more context information, will be the main research direction of the future.

Nevertheless, it’s still too early to be optimistic. \cite{jia_adversarial_2017} tested whether systems can answer questions about paragraphs that contain adversarially inserted sentences, which are automatically generated to distract computer systems without changing the correct answer or misleading humans. The result shows that the accuracy of sixteen published models drops from an average of 75\% F1 score to 36\%. \cite{sugawara_what_2018} also employed simple heuristics to split each dataset into easy and hard subsets and the experiment showed that the existing models perform worse in the latter, where the questions need more reasoning skills. Therefore, the efficient and accurate language models (such as BERT, GPT2.0) are necessary to help the machine learn reasoning. Although BERT has a visible improvement for NLP tasks, but it requires a lot of computing resources and training time, which is not something that the average team can afford. Whether it is possible to find a more efficient method and to train an efficient language model with a small amount of resources are essential to further study.

\section{Conclusion}

We have surveyed the main datasets and models in the MRC task in recent years. In terms of datasets, the span-prediction and open-form styles have been more popular. As for models, the improvements are mainly made on the Match Layer. The proposed pre-training model recently has achieved State-Of-The-Art performance on several datasets and is the key research direction in the future MRC field.

\appendix

\bibliographystyle{named}
\bibliography{ijcai19}

\end{document}